\documentclass[runningheads]{llncs}
\usepackage[T1]{fontenc}
\usepackage{graphicx}
\usepackage{amsmath}
\usepackage{amssymb}
\usepackage{bm}
\usepackage{multirow}
\usepackage{booktabs}
\usepackage{makecell}
\usepackage{marvosym}
\begin{document}
\raggedbottom
\title{Vision-Language-Action Model Optimization for Real-Time Robotic Manipulation}
\titlerunning{VLA Model Optimization for Real-Time Robotic Manipulation}
%

\author{Juyi Lin\inst{1}\and
Amir Taherin\inst{1} \and
Arash Akbari\inst{1} \and
Arman Akbari\inst{1} \and
Lei Lu\inst{1} \and
Guangyu Chen\inst{1} \and
Taskin Padir\inst{1} \and
Xiaomeng Yang\inst{1} \and
Jayanth Srinivasa\inst{2} \and
Xue Lin\inst{1} \and
David Kaeli\inst{1} \and
Yanzhi Wang\inst{1,3} \and
Pu Zhao\inst{1}\textsuperscript{(\Letter)} \and
Xuan Shen\inst{1}\textsuperscript{(\Letter)}}
\authorrunning{J. Lin et al.}
%
\institute{Northeastern University, Boston, USA\\
 \and
Cisco, Chicago, USA \\
\and
EmbodyX,San Mateo, USA \\
\email{\{lin.juy, p.zhao, shen.xuan\}@northeastern.edu}
}


%
\maketitle              
\begin{abstract}
Recent large-scale Vision Language Action (VLA) models have shown superior performance in robotic manipulation tasks guided by natural language. 
However, current VLA models suffer from two drawbacks: (i)  generation of massive tokens leading to high inference latency and increased training cost, and (ii) insufficient utilization of generated actions resulting in potential performance loss. 
To address these issues, we develop a training framework to finetune VLA models for generating significantly fewer action tokens with high parallelism, effectively reducing inference latency and training cost. 
Furthermore, we introduce an inference optimization technique with a voting ensemble strategy to combine current and previous action predictions, improving the utilization of generated actions and overall performance. 
Our results demonstrate that we achieve superior performance, achieving significantly higher success rates and 39$\times$ faster inference than OpenVLA with 46 Hz throughput on edge platforms, demonstrating practical deployability. The code is available at \url{https://github.com/LukeLIN-web/VOTE}.

\end{abstract}

\section{Introduction}
\label{sec:introduction}

\begin{figure*}[t]
\centering
\includegraphics[width=0.8\textwidth]{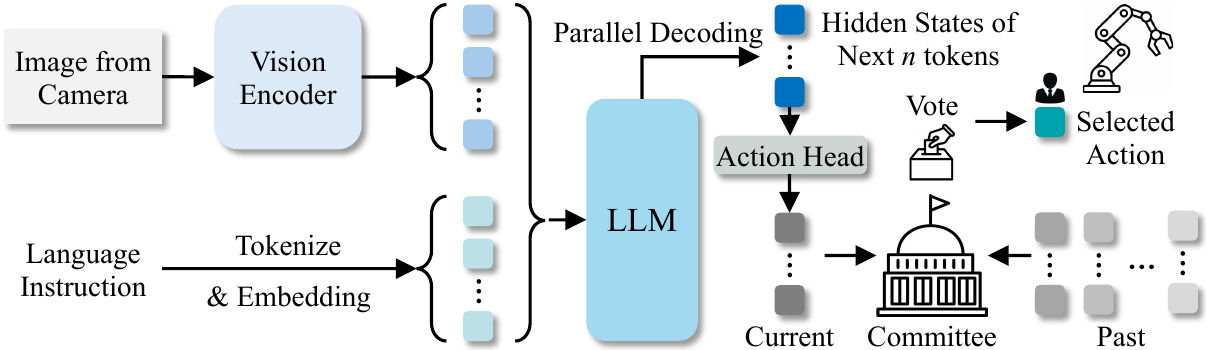}
\caption{The whole VOTE pipeline, where we generate the next following $n$ actions in parallel and adopt the ensemble voting strategy for  accurate  action prediction.}
\label{fig:whole_pipeline}
\vspace{-10pt}
\end{figure*}

Building general-purpose robotic policies capable of handling diverse tasks, embodiments, and real-world interactions has been a central challenge in robotics research.
Recent studies~\cite{kim2024openvla,zhu2025objectvla,qu2025spatialvla,li2024cogact} leverage Vision-Language-Action (VLA) models to address this problem, demonstrating excellent accuracy across a variety of robotic tasks.
VLA models enable robots to perform complex tasks from natural language instructions, achieving outstanding performance on familiar objects and environments within the training distribution~\cite{brohan2023rt2visionlanguageactionmodelstransfer,o2024open,kim2025fine,li2023vision}.
The VLA models are mainly developed based on  Vision Language Models (VLMs)~\cite{chen2023pali,driess2023palm,karamcheti2024prismatic,zhao2025open,beyer2024paligemma} 
through continuous training or finetuning  on diverse robot data~\cite{o2024open,fang2024rh20t}.
The success of this paradigm lies in leveraging the generalization capabilities of VLMs across diverse robotic manipulation tasks, alongside architectural designs that effectively integrate the VLM backbone with the robot action output head.

However, the slow inference  of VLA models significantly hinders their practical deployment. 
This latency challenge is often aggravated by performance-enhancing modifications \cite{mi2026effective,shen2025fastcar,shen2025draftattention,shen2025efficient,shen2025sparse,zhan2024exploring,zhan-etal-2024-rethinking-token}. 
For example, SpatialVLA incorporate additional visual modules for 3D position encoding, 
yielding superior evaluation results.
However, reliance on additional visual features introduces significant pre-processing overhead, and the increased number of visual tokens results in longer input sequences, further impacting inference speed.
The high training cost across diverse robotic datasets, and the  latency of action sampling, limits the scalability and practical deployment of VLA models in real-world scenarios \cite{zhan2024fast,shen2024lazydit,shen2024numerical,zhao-etal-2024-pruning,shen2024search,yang2023pruning,shen2025quartdepth,yang2026alter}. This motivates us to investigate fast-acting efficient methods to reduce training and inference overhead.

We identify two main problems in existing VLA models: (i) generation of massive tokens leading to large inference latency and more training cost, and (ii) insufficient utilization of generated actions resulting in potential performance loss. To address these problems, we propose \textbf{VOTE}, a lightweight VLA training method as shown in Figure~\ref{fig:whole_pipeline} to generate much fewer  tokens for superior efficiency, and a plug-and-play ensemble voting strategy for accuracy enhancement.

Specifically, we introduce a special token \texttt{<ACT>} to represent entire action chunks. Thus,  the model only needs to generate one single \texttt{<ACT>} token, instead of the original multiple tokens,  significantly reducing the number of generated tokens and avoiding multiple sequential decoder passes with tokenization. 
Consequently, it enables faster inference and substantially reduces training cost, making rapid fine-tuning practical due to fewer required input-output tokens and simpler decoding.
Meanwhile, we propose a novel action-ensemble technique, \emph{ensemble voting}, to improve the model performance at test time.
Specifically, we construct an action ensemble committee by incorporating actions predicted in previous steps, and determine the current action based on a voting mechanism weighted by the accumulated “tickets” from prior predictions.
Experimental results demonstrate that our method achieves superior performance, while also delivering higher throughput and faster inference.
We improve the average success rate (ASR) of OpenVLA by over 20\% across four LIBERO task suites, surpass CogACT by 7\% ASR on SimplerEnv WidowX Robot, and accelerate action generation throughput by 39$\times$ on the edge device NVIDIA Jetson Orin.

Our contributions are summarized as follows:
\begin{enumerate}
    \item 
    We propose VOTE, including a training framework and an inference optimization technique. With the training framework, the VLA model only needs to generate one single special token instead of multiple tokens, thus effectively reducing the training and inference costs, while further improving the action success rate. 

    \item For  inference, we propose a novel action ensemble technique to construct a committee for action selection with voting,  improving  action success rate.
    
    \item Experiments show that VOTE achieves high success rate, with lower training costs and significantly improved inference speedups with higher throughput.

\end{enumerate}

\section{Related works and motivation}
\label{sec:motivation}

Bridging the gap between seeing, understanding, and acting, Vision-Language-Action (VLA) models represent a significant leap in robotics and object manipulation. More recently, several studies~\cite{brohan2022rt,zitkovich2023rt,shentu2024llms,li2024cogact,qu2025spatialvla,black2024pi_0,li2024towards,li2025hamster,taherin2025cross} present new ways of building general robot policies by fine-tuning pretrained VLMs on robot data, offering the ability to directly generate robot actions. OpenVLA \cite{kim2024openvla} proposes to fine-tune the prismatic VLM~\cite{karamcheti2024prismatic} only on the OXE dataset~\cite{o2024open}. CogACT~\cite{li2024cogact} employs a diffusion transformer-based action module to enhance generalization and adaptability in robotic tasks. SpatialVLA~\cite{qu2025spatialvla} introduces Ego3D Position Encoding to inject 3D information into the input observations of their VLA, representing spatial robot movement actions with Adaptive Action Grids. 

We point out two major issues in existing VLAs detailed below: (i) the production of large token sequences, and (ii) the limited use of generated actions. 

\subsection{Massive Computational Overhead}

Typical VLA models need to predict multiple tokens corresponding to different action dimensions for each action, leading to high inference latency and training cost, whereas diffusion-based VLAs also introduce additional computational overhead from more training steps and multi-step denoising, as specified below. 

\begin{figure}[t]
    \centering
    \includegraphics[width=0.8\linewidth]{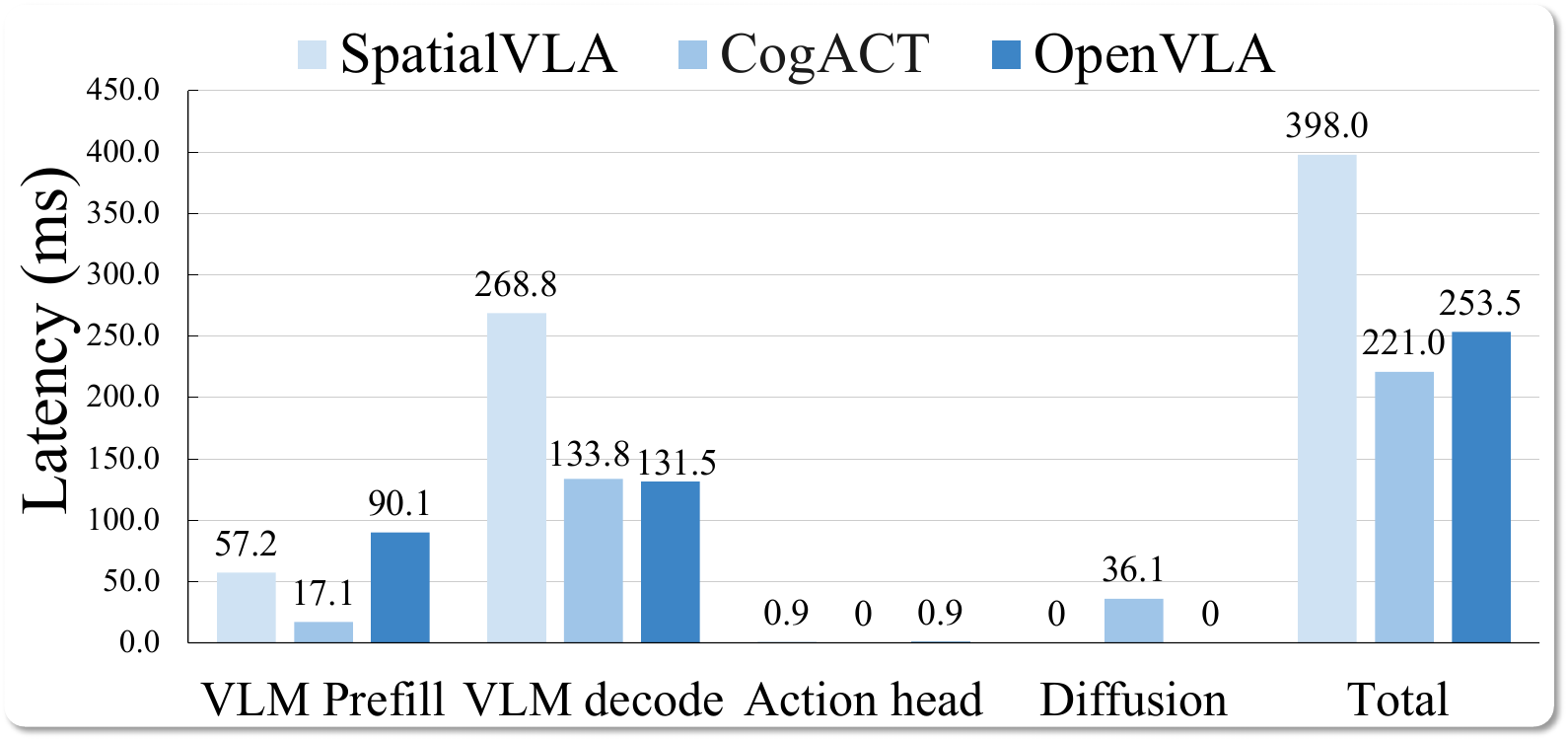}
    \caption{Latency for SpatialVLA, CogACT, and OpenVLA.} 
    \label{fig:combined_metrics}
    \vspace{-10pt}
\end{figure}

\textbf{Large Inference Latency.} We show the latency profile for SpatialVLA, OpenVLA and CogACT, in Figure~\ref{fig:combined_metrics}. As observed, the primary computational overhead in current VLA models lies in the VLM backbone within the VLA architecture. The VLM decoding, which needs to generate large amounts of tokens, dominates the overall latency for action prediction with at least 50\% occupation across three models.
In particular, the diffusion  in CogACT incurs additional latency overhead. 
Meanwhile, SpatialVLA relies on multimodal high-level visual representations, such as 3D information, which needs to feed massive additional visual input tokens to the VLM with significantly increased latency. 

\textbf{Massive Training Cost.}
VLA models normally adopt finetuning with data from new tasks and embodiments to improve the performance in new environments~\cite{li2024cogact}.
Existing methods rely on large-scale pretraining data and additional downstream data (such as Fractal~\cite{brohan2022rt} and BridgeDataV2~\cite{walke2023bridgedata})  to adapt VLM backbones for robotic action prediction tasks.
Moreover, during training, as we need to pad multiple empty action embeddings (corresponding to the number of output tokens for actions) as inputs, a  large number of output tokens leads to padding massive input tokens, incurring significant additional training cost along with massive training data.
OpenVLA-OFT~\cite{kim2025fine} shows that the diffusion action head converges more slowly and requires 1.67$\times$–2$\times$ more gradient steps to converge compared to the MLP action head.

\subsection{Insufficient Utilization of Generated Actions}

Although VLA models generate large amounts of actions at high training and inference costs, we note that not all generated actions are utilized effectively. At each time step, the VLA model predicts a sequence of actions for the next multiple time steps. Thus, at each  step, the robot receives the action from the current inference, as well as the historical predictions from previous model inferences. 

Typically,  robots directly execute the action from the current inference based on the current observation, 
discarding historical action predictions of previous time steps. This approach fails to fully utilize historical visual information and model predictions, leading to a less stable trajectory with potential performance degradation. 
To address this, prior works \cite{zhao2023learning,li2024cogact} propose to combine actions predicted for the current time step from both present and past inferences. 
However, these methods suffer from ineffective combination with meaningless outputs, too simple combination strategy, or heavy reliance on the current prediction which could be incorrect. Actions predicted at different timesteps can belong to different modes~\cite{chi2023diffusion}, and simply aggregating them could result in an action that does not align with any modes. 
The utilization of generated actions from current and past inferences are insufficient, resulting in degraded performance.

\section{Method}
\label{sec:our_method}

Motivated by the above drawbacks of the current VLA models, we develop a training framework to finetune VLA models for generating less action tokens, thus addressing the first drawback with reduced inference latency and training cost. Furthermore, an inference optimization is proposed with a novel ensemble strategy to combine actions of current and previous predictions, thus addressing the second drawback with improved utilization of generations.

In this section, we first briefly provide the preliminaries of our model's architecture and problem statement. 
We next elaborate our innovative training method, detailing how the introduction of the special token \texttt{<ACT>} optimizes action prediction accuracy and computational efficiency. 
Then we introduce our vote-based adaptive action ensemble strategy, designed to enhance the stability and robustness of action execution by dynamically selecting relevant actions.

\subsection{Problem Statement}
\label{sec:Problemstatement}
Our model generates an action based on the image $\bm I \in \mathbb{R}^{W \times H \times 3}$ and the language instruction $\bm{l}$. At time step $t$, we utilize a model $\pi$ to predict a temporal action sequence $(\bm{a}_t,\bm{a}_{t+1},...,\bm{a}_{t+N})$ for executing the desired task:
\begin{equation}
\small
    \bm{\pi}: (\bm{l},\bm{I}_t) \rightarrow (\bm{a}_t,\bm{a}_{t+1},...,\bm{a}_{t+N}) \label{eq:target}
\end{equation}

Here, $\bm{a}_t$ can describe various robot actions with different control modes and end-effectors. Following a strategy described in previous work~\cite{kim2024openvla,kim2025fine,li2024cogact}, we use 7 degrees of freedom (DoF)
to express the end-effector pose of the robot arm:
\begin{equation}
\small
    \bm{a}_t = [\Delta x, \Delta y, \Delta z, \Delta \phi, \Delta \theta, \Delta \psi, g] \label{eq:action}
\end{equation}
where $\Delta x, \Delta y, \Delta z$ are the relative translation offsets of the end effector, $\Delta \phi, \Delta \theta, \Delta \psi$ denote the rotation changes, and $g \in \{0,1\}$ indicates the gripper’s open/close state. This action space enables continuous control over robot arm motion.

\subsection{Training Framework}
\label{sec:trainingmethod}

\textbf{Overview.} During training, we introduce a special token \texttt{<ACT>} into the tokenizer of the LLM to explicitly signal the action prediction task. Specifically, we append this \texttt{<ACT>} token to the end of each language instruction sequence as the target token label. After the LLM performs a single forward pass to generates the single token, its hidden state from the final-layer is passed to the action head for transformation into  continuous action values $\hat{\bm{a}}$.  We specify the details of our training framework below. 

\textbf{Action Generation.} Given a language instruction $\bm{l}$ and corresponding image $\bm I$, the model generates multiple consecutive action predictions. 
First, the input data is processed by the VLA model to obtain hidden states:
\begin{equation}
\label{eq:eq3}
\small
\bm{h} = \text{VLA}(\bm{l}, I), \quad \text{where}\quad \bm{h}\in \mathbb{R}^{B \times \text{L}\times H},
\end{equation}
where $B$ is the batch size, $L$ is the sequence length, and $H$ is  hidden dimension.

Next,  hidden states corresponding to the special  token  \texttt{<ACT>} are extracted:
\begin{equation}
\small
\bm{h}_{\texttt{<ACT>}} = \bm{h}[\text{mask}_{\texttt{<ACT>}}],\quad \text{where} \quad \bm{h}_{\texttt{<ACT>}} \in \mathbb{R}^{B \times 1 \times H}.
\end{equation}
Note that model only needs to generate one single token $\texttt{<ACT>}$ instead of multiple tokens for various multi-dimensional actions.  

Then we need to convert the hidden state of $\texttt{<ACT>}$ to actual action predictions. This is achieved with an action head. Specifically, 
the  hidden state $\bm{h}_{\texttt{<ACT>}}$ is  passed through an MLP Action Head to predict the action chunk (multiple consecutive actions).
The Action Head first downsamples the dimension of $\bm{h}_{\texttt{<ACT>}}$ from 4096 to 1024, then passes it through 4 bottleneck blocks. Each bottleneck block upsamples the dimension back to 4096, then downsamples it to 1024 again. The block architecture is shown below:

\begin{equation}
\small
\bm{out} = \bm{x} + \operatorname{Dropout}(\operatorname{SiLU}\bigl(\operatorname{LayerNorm}(\bm{x})\bm{W}_\mathrm{up}\bigr) \bm{W}_\mathrm{down})
\end{equation}

This bottleneck design achieves better performance with fewer parameters compared with the isotropic architecture where all linear layers have the same dimension as $\bm{h}_{\texttt{<ACT>}}$.

\textbf{Training Objectives.} Our training objective incorporates both token-level and action-level supervision. 

We use $\bm{L}_{\mathrm{action}}$ to represent the L1 loss between the predicted actions $\hat{\bm{a}}$  and  the ground-truth actions $\bm{a}$
across all action dimensions. 
\begin{align}
\small
\begin{aligned}
\bm{L}_{\mathrm{action}} = \bm{L_1}(\hat{\bm{a}}, \bm{a})=\frac{1}{BNA}\sum_{b=1}^{B}\sum_{n=1}^{N}\|\hat{\bm{a}}_{b,n}-\bm{a}_{b,n}\|_1 ,
\end{aligned}
\end{align}

Meanwhile, $\bm{L}_{\mathrm{token}}$ is the cross-entropy loss calculated based on the prediction of the $\texttt{<ACT>}$ token and all instruction tokens. These two losses are combined into a weighted total loss that balances semantic understanding from language modeling and accurate action generation:
\begin{equation}
\small
\bm{L}_{\mathrm{total}} = \lambda_{\mathrm{token}} \bm{L}_{\mathrm{token}} + \lambda_{\mathrm{action}} \bm{L}_{\mathrm{action}},
\end{equation}

\textbf{Advantages for Training and Inference.} Our method condenses the entire action chunk into a compact, high-level representation using a single $\texttt{<ACT>}$ token. The hidden state of this token is passed through the action head to directly predict all action chunks. This significantly reduces the number of tokens required, leading to improved efficiency in both training and inference.

Furthermore, instead of employing an action tokenizer to convert action tokens into actual actions, we directly utilize an action head to map the hidden state of the special token \texttt{<ACT>} to normalized continuous actions, enabling efficient parallel computation and eliminating the need for action tokenizer. 

\subsection{Ensemble Voting}
\label{sec:vote}

\begin{figure*}[!tpb]
    \centering
    \includegraphics[width=0.9\linewidth]{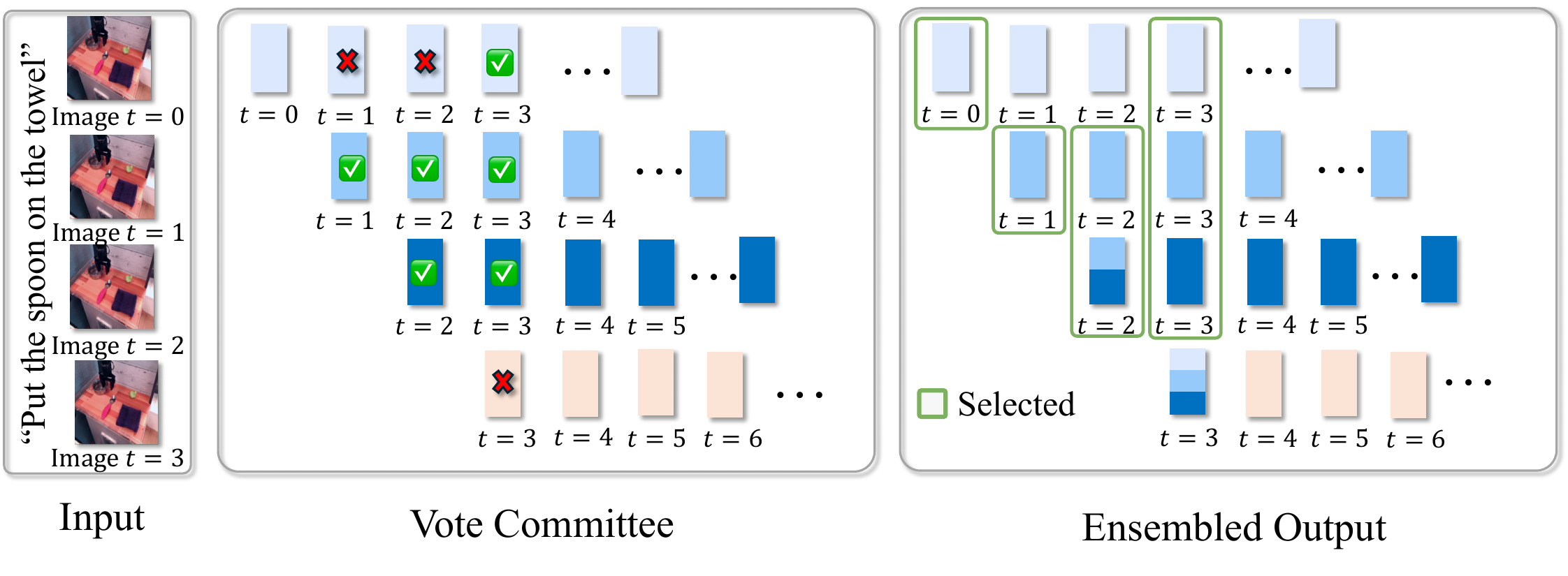}
    \caption{\textbf{Vote Action Ensemble.} Illustration of our action ensemble strategy with K = 3 (using the last 3 historical action predictions) as an example. Historical predictions and the current prediction form a voting committee to jointly determine the final action to execute. For example, when t = 3, more than half of the candidate actions differ from the current prediction, voting not-similar. Therefore, we discard the current prediction and instead compute the final ensembled action by averaging the previous 3 historical action predictions which vote not-similar.}  
    \label{fig:ensemble}
    \vspace{-10pt}
\end{figure*}

During inference, the VLA model predicts a sequence of actions across multiple time steps. Typically, the robots execute these actions consecutively based on the current observation, discarding historical action predictions of previous time steps. However, this approach fails to fully utilize historical visual information and model predictions, leading to a less stable trajectory with potential performance degradation. 

To fully utilize the generated actions, we propose a voting-based adaptive ensemble strategy for action aggregation, which  selects the more frequent prediction (with a higher chance to be correct) from a  list  of action predictions from adjacent time steps, as illustrated in Figure ~\ref{fig:ensemble}.
Specifically, given the current observation $\bm{o}_t$, let $(\bm{a}_t|\bm{o}_t)$ denote the predicted action at current time step $t$. Note that one inference can generate multiple consecutive actions, i.e., $(\bm{a}_t|\bm{o}_t)$, $(\bm{a}_{t+1}|\bm{o}_t)$, $(\bm{a}_{t+2}|\bm{o}_t)$, and so on.
At the current time step $t$, the past action predictions from previous time steps are also available, represented as:   $\bm{H}=\{(\bm{a}_t|\bm{o}_{t-K}), ..., (\bm{a}_t|\bm{o}_{t})\}$.

We first compute the similarity between each action in $\bm{H}$ and the current/newest prediction $(\bm{a}_t|\bm{o}_{t})$. 
Based on all these similarities, the action set $\bm{H}$ is split into two subsets, $\bm{M}$ for higher similarity and $\bm{N}$ for lower similarity. Following the common voting rule, we select the set with more votes and compute the average of all actions in the selected subset as the final action for the time step $t$.
The ensemble action $\tilde{\bm{a}}_t$  at time step $t$ is computed by the following:
\begin{equation}
\label{eq:vote_ensemble}
\small
\hat{\bm{a}}_t = 
\begin{cases}
\displaystyle\frac{1}{|\bm{M}|} \sum\limits_{\bm x \in \bm{M}} \bm x, & \text{if } |\bm{M}| > |\bm{N}|, \\
\\[-1ex]
\displaystyle\frac{1}{|\bm{N}|} \sum\limits_{\bm x \in \bm{N}} \bm x, & \text{otherwise,}
\end{cases}
\end{equation}
\begin{equation}
\small
\bm{M} = \left\{ (\bm{a}_t|\bm{o}_{t-k}) \;\middle|\; \langle \bm{a}_t|\bm{o}_t, \bm{a}_t|\bm{o}_{t-k} \rangle > \tau,\; k \in \{0, \dots, K\} \right\},
\end{equation}
\begin{equation}
\small
\bm{N} = \left\{ (\bm{a}_t|\bm{o}_{t-k}) \;\middle|\; \langle \bm{a}_t|\bm{o}_t, \bm{a}_t|\bm{o}_{t-k} \rangle \le \tau,\; k \in \{0, \dots, K\} \right\},
\end{equation}
where $\langle \cdot, \cdot \rangle$ denotes cosine similarity, $\tau$ is a threshold empirically set to $0.5$, and $|\cdot|$ is the element number in a set.

\textbf{Advantages.}   
(i)~Unlike straightforward action chunking, where actions are executed consecutively, or static weighted aggregation methods in \cite{zhao2023learning}, our method selects the actions with more votes, with higher probability to be correct. 
(ii) Unlike naive averaging or the adaptive method~\cite{li2024cogact}, our ensemble voting filters out inconsistent mode predictions, enhancing the robustness of the final action.
(iii) Our method pays more attention or gives more credit to the current/newest action prediction, since all similarities are computed with reference to the current action with one definite vote for high similarity.  This is reasonable as the current observation provides the most critical real-time information.

\begin{table}[t]
\centering
\caption{Evaluation on SimplerEnv WidowX and LIBERO benchmarks. SimplerEnv results are averaged over all four tasks; latency is tested on an A6000 GPU. LIBERO reports SR across task suites (Chunk=8). Ours achieves the best performance.}
\label{tab:widowx_libero}
\vspace{-6pt}
\resizebox{\textwidth}{!}{%
\setlength\tabcolsep{3pt}
\begin{tabular}{l|c|c|c|c|c|c|c|c}
\toprule
\multirow{2}{*}{Method} & \multicolumn{3}{c|}{SimplerEnv WidowX} & \multicolumn{5}{c}{LIBERO} \\
\cmidrule(lr){2-4} \cmidrule(lr){5-9}
 & Avg.(\%) & Lat.(ms)$\downarrow$ & Spd.$\uparrow$ & Spatial & Object & Goal & Long & Avg. \\
 &  &  &  & SR(\%) & SR(\%) & SR(\%) & SR(\%) & SR(\%) \\
\midrule
RT-1-X \cite{o2024open} & 1.1 & -- & -- & -- & -- & -- & -- & -- \\
Octo-Base \cite{team2024octo} & 16.0 & -- & -- & -- & -- & -- & -- & -- \\
Octo-Small \cite{team2024octo} & 30.0 & -- & -- & -- & -- & -- & -- & -- \\
OpenVLA \cite{kim2024openvla} & 1.0 & 240 & 1.0 & 84.7 & 88.4 & 79.2 & 53.7 & 76.5 \\
RoboVLM \cite{li2024towards} & 31.3 & -- & -- & -- & -- & -- & -- & -- \\
Diffusion Policy \cite{chi2023diffusion} & -- & -- & -- & 78.3 & 92.5 & 68.3 & 50.5 & 72.4 \\
Octo \cite{team2024octo} & -- & -- & -- & 78.9 & 85.7 & 84.6 & 51.1 & 75.1 \\
TraceVLA \cite{zheng2024tracevla} & -- & -- & -- & 84.6 & 85.2 & 75.1 & 54.1 & 74.8 \\
$\pi_0$ \cite{black2024pi_0} & 27.1 & -- & -- & 96.8 & 98.8 & 95.8 & 85.2 & 94.2 \\
$\pi_0$-FAST \cite{pertsch2025fast} & 32.1 & 470 & 0.5 & 96.4 & 96.8 & 88.6 & 60.2 & 85.5 \\
SpatialVLA \cite{qu2025spatialvla} & 42.7 & 400 & 0.6 & 88.2 & 89.9 & 78.6 & 55.5 & 78.1 \\
CogACT \cite{li2024cogact} & 51.3 & 220 & 1.1 & -- & -- & -- & -- & -- \\
OpenVLA-OFT \cite{kim2025fine} & -- & -- & -- & 96.2 & 98.3 & 96.2 & 90.7 & 95.3 \\
\midrule
Ours & \textbf{58.3} & \textbf{78} & \textbf{3.1} & \textbf{98.8} & \textbf{99.8} & \textbf{97.6} & \textbf{95.6} & \textbf{98.0} \\
\bottomrule
\end{tabular}}%
\end{table}

\begin{table*}[t]
\centering
\caption{\textbf{Cross-Platform Inference Evaluation}. Peak VRAM represents the maximum GPU memory used during inference. Speedup is reported relative to OpenVLA as the baseline. Bold denotes the best performance for each metrics.}
\label{tab:inference_evaluation}
\vspace{-6pt}
\setlength\tabcolsep{3pt}
\begin{tabular}{@{}lc|cccc|cccc@{}}
\toprule
\multicolumn{1}{c}{\multirow{2}{*}{Model}} &
  \multirow{2}{*}{\begin{tabular}[c]{@{}c@{}}Chunk\\Size\end{tabular}} &
  \multicolumn{4}{c|}{\textbf{NVIDIA RTX A6000}} &
  \multicolumn{4}{c}{\textbf{NVIDIA Jetson Orin}} \\
& &
  {\begin{tabular}[c]{@{}c@{}}Latency\\(ms)~$\downarrow$\end{tabular}} &
  {\begin{tabular}[c]{@{}c@{}}VRAM\\(GB)~$\downarrow$\end{tabular}} &
  {\begin{tabular}[c]{@{}c@{}}Throu.\\(Hz)~$\uparrow$\end{tabular}} &
  {\begin{tabular}[c]{@{}c@{}}Speed\\up~$\uparrow$\end{tabular}} &
  {\begin{tabular}[c]{@{}c@{}}Latency\\(ms)~$\downarrow$\end{tabular}} &
  {\begin{tabular}[c]{@{}c@{}}VRAM\\(GB)~$\downarrow$\end{tabular}} &
  {\begin{tabular}[c]{@{}c@{}}Throu.\\(Hz)~$\uparrow$\end{tabular}} &
  {\begin{tabular}[c]{@{}c@{}}Speed\\up~$\uparrow$\end{tabular}} \\
\midrule
OpenVLA       & 1  & 240  & 14.35 & 4.2   & 1.0  & 836  & 14.35 & 1.2   & 1.0  \\
SpatialVLA    & 4  & 400  & 7.82  & 10.1  & 2.4  & 1949 & 7.82  & 2.1   & 1.7  \\
CogACT        & 16 & 220  & 29.33 & 72.4  & 17.7 & --   & OOM   & --    & --   \\
\midrule
Ours          & 8  & 78   & 14.40 & 102.6 & 24.4 & 346  & 14.40 & 23.1  & 19.3 \\
Ours          & 16 & 78   & 14.40 & \textbf{205.2} & \textbf{48.8} & 346  & 14.40 & \textbf{46.2}  & \textbf{38.6} \\
\bottomrule
\end{tabular}
\end{table*}

\section{Experimental Results}

\subsection{Experimental Setup and Baselines}
\label{sec:exp_setups}

We evaluate our model on the LIBERO~\cite{liu2023libero} and SimplerEnv~\cite{li24simpler} simulation benchmarks, which comprise a diverse set of robotic manipulation tasks in simulated environments. We fine-tune on OpenVLA using AdamW with a learning rate of $1 \times 10^{-4}$ and LoRA~\cite{hu2022lora} with rank $r=32$ and $\alpha=16$. The action chunk size $N$ and ensemble horizon $K+1$ (including the current prediction) are both set to 8. For LIBERO, we train on 2 H100 GPUs with a global batch size of 40. For SimplerEnv, we train on 4 H100 GPUs with a global batch size of 80, using the same datasets as \cite{qu2025spatialvla}, with 60K steps on BridgeDataV2~\cite{walke2023bridgedata} for \textbf{WidowX robot} and 70K steps on Fractal for \textbf{Google robot}.

\subsection{Performance Evaluation}

Table~\ref{tab:widowx_libero} summarizes results on SimplerEnv~\cite{li24simpler} and LIBERO. On the WidowX setup (24 trials per task), our model surpasses CogACT~\cite{li2024cogact} and SpatialVLA with 58.3\% average success rate and over 3$\times$ speedup over OpenVLA. On LIBERO with chunk size 8, our method performs best in all sub-tasks, demonstrating that a single \texttt{<ACT>} token has better representation ability than multiple action tokens.

\begin{table}[t]
\centering
\caption{Comparison on LIBERO across different architectures and chunk sizes.}
\label{tab:variant_libero}
\vspace{-6pt}
\setlength\tabcolsep{1pt}
\begin{tabular}{l|c|c|c|c|c}
\hline
\multirow{2}{*}{\makecell{Models}} & Spatial & Object & Goal & Long & Average \\
& SR (\%) & SR (\%) & SR (\%) & SR (\%) & SR (\%) \\
\hline
Isotropic (Chunk=8) & 98.0 & 99.5 & 96.0 & 94.0 & 96.9 \\
Isotropic (Chunk=16)  & 96.0 & 98.5 & 94.0 & 91.0 & 94.9 \\
Bottleneck (Chunk=8) & \textbf{98.8} & \textbf{99.8} & \textbf{97.6} & \textbf{95.6} & \textbf{98.0} \\
Bottleneck (Chunk=16) & 97.6 & 97.8 & 96.8 & 93.8 & 96.5 \\
\hline
\end{tabular}
\end{table}

\subsection{Cross-Platform Inference Evaluation}
\label{sec:orin_evaluation}

To investigate the efficiency of \textbf{VOTE}, we measured the average latency (i.e., the time to generate an action chunk) and throughput (i.e., the number of actions generated per second) by querying each model 100 times. Each query processes a 224$\times$224 image and a sample language instruction (``\textit{What action should the robot take to pick the cup?}'').

We first test the inference throughput on the A6000 GPU. As shown in Table~\ref{tab:inference_evaluation}, \textbf{VOTE} achieves a throughput of approximately $20\times$ of SpatialVLA~\cite{qu2025spatialvla}, despite the larger LLM used in our model (our LLaMA2-7B versus SpatialVLA's PaliGemma-3B). With chunk 16, \textbf{VOTE} can deliver up to 48.8$\times$ speed up compared to OpenVLA, outperforming other baselines. 
The difference in speedups compared to Table~\ref{tab:widowx_libero} arises from different evaluation setting. In Table~\ref{tab:widowx_libero}, for difficult tasks such as SimplerEnv, we use the same setting as CogACT and SpatialVLA to predict one action chunk at each timestep without finishing executing all actions in this chunk.
But in Table~\ref{tab:inference_evaluation}, for easier tasks such as LIBERO, we predict the next action chunk only after all actions in the chunk have been executed. Our method achieves high throughput on easier tasks while achieves lower latency in scenarios that demand higher manipulation precision.

Meanwhile, modern edge-computing platforms, such as the NVIDIA AGX Orin~\cite{Orin_specification}, are preferred for real-time robotic control, enabling real-time robot inference. 
However, these platforms suffer from 
the heavy demands of VLA models due to limited and heterogeneous computing resources. To assess performance on the edge platform NVIDIA Jetson Orin, we compare our proposed approach with existing methods including OpenVLA~\cite{kim2024openvla}, SpatialVLA~\cite{qu2025spatialvla}, CogACT~\cite{li2024cogact}. As shown in Table~\ref{tab:inference_evaluation}, \textbf{VOTE} (with a chunk size  16) achieves 46~Hz throughput and a 38.6$\times$ speedup over OpenVLA, whereas CogACT fails to execute due to Out-of-Memory (OOM). These results highlight our superior latency and throughput, making \textbf{VOTE} well-suited for edge deployment.

\begin{table}[t]
  \centering
  \caption{Comparison of our proposed action ensemble strategy, Ensemble Voting, with other strategies. The Average strategy simply averages all historical predictions.}
  \label{tab:voteabalation}
  \vspace{-6pt}
  \setlength\tabcolsep{2pt}
  \begin{tabular}{c|c|c|c|c}
    \hline
   \multirow{2}{*}{ Strategy} & GR VM & GA VA & WR & Average \\
    & SR (\%) & SR (\%) & SR (\%) & SR (\%) \\
    \hline
None     & 60.9 & 56.3 & 24.0 & 47.1 \\
Temporal \cite{zhao2023learning}   & 70.5 & 60.3 & 30.2 & 53.7 \\
Adaptive \cite{li2024cogact} & 71.9 & 60.3 & 49.0 & 60.4 \\
Average      & 72.1   & 59.6 & 53.2 &  61.6 \\
Ensemble Voting   & \textbf{74.9} & 60.2 & \textbf{58.3} & \textbf{64.5} \\
    \hline
  \end{tabular}
\end{table}

\subsection{Ablation Study}

We conduct ablation studies on SimplerEnv across two robot embodiments: Google Robot (GR) and WidowX Robot (WR). We use the following abbreviations: VM for   SimplerEnv Visual Matching setting, and VA for   SimplerEnv Visual Aggregation setting.
We analyze the impact of chunk size and model architecture. 
Table~\ref{tab:variant_libero} compares models with chunk sizes of 8 and 16 under both isotropic and bottleneck architectures for the action head. 
We observe that the bottleneck architecture consistently outperforms the isotropic one across both chunk sizes. 
Our model maintains superior performance when predicting 16 subsequent actions from a single observation, exhibiting only a modest 1.5\% reduction in average success rate compared to a chunk size of 8.
We ablate the effect of the ensemble voting strategy in Table~\ref{tab:voteabalation}. Our method outperforms all other strategies and is also robust to hyperparameter $\tau$, achieving optimal performance of 58.3\% at $\tau = 0.5$ and outperforming other strategies across $\tau \in [0.0, 0.8]$.

\section{Conclusion}

We have presented a lightweight VLA framework that enhances efficiency by predicting actions in a hidden latent space. Our approach leverages an action-tokenizer-free training methodology that simultaneously predicts multiple actions with bottleneck action head, significantly reducing computational requirements during both training and inference. 
Furthermore, we propose a straightforward yet effective action ensemble algorithm that optimizes action sampling. Extensive experimental results confirm that our model achieves superior inference speedups, while exhibiting exceptional generative performance.

%
%
%
\small
\bibliographystyle{splncs04}
\bibliography{main}

@article{kim2024openvla,
  title={Openvla: An open-source vision-language-action model},
  author={Kim, Moo Jin and Pertsch, Karl and Karamcheti, Siddharth and   others},
  journal={arXiv preprint arXiv:2406.09246},
  year={2024}
}

@article{liu2023libero,
  title={Libero: Benchmarking knowledge transfer for lifelong robot learning},
  author={Liu, Bo and Zhu, Yifeng and Gao, Chongkai and others},
  journal={NeurIPS},
  year={2023}
}

@inproceedings{zitkovich2023rt,
  title={Rt-2: Vision-language-action models transfer web knowledge to robotic control},
  author={Zitkovich, Brianna and Yu, Tianhe and  others},
  booktitle={Conference on Robot Learning},
  year={2023},
  organization={PMLR}
}

@article{li24simpler,
         title={Evaluating Real-World Robot Manipulation Policies in Simulation},
         author={Xuanlin Li and Kyle Hsu and Jiayuan Gu and others},
         journal = {arXiv preprint arXiv:2405.05941},
         year={2024}
}

@article{qu2025spatialvla,
  title={SpatialVLA: Exploring Spatial Representations for Visual-Language-Action Model},
  author={Qu, Delin and Song, Haoming and Chen, Qizhi and  others},
  journal={arXiv preprint arXiv:2501.15830},
  year={2025}
}

@article{zhao2023learning,
  title={Learning fine-grained bimanual manipulation with low-cost hardware},
  author={Zhao, Tony Z and Kumar, Vikash and Levine, Sergey and Finn, Chelsea},
  journal={arXiv preprint arXiv:2304.13705},
  year={2023}
}

@inproceedings{fang2024rh20t,
  title={Rh20t: A comprehensive robotic dataset for learning diverse skills in one-shot},
  author={Fang, Hao-Shu and Fang, Hongjie and Tang, Zhenyu and others},
  booktitle={ICRA},
  pages={653--660},
  year={2024},
  organization={IEEE}
}

@article{chi2023diffusion,
  title={Diffusion policy: Visuomotor policy learning via action diffusion},
  author={Chi, Cheng and Xu, Zhenjia and Feng, Siyuan and others},
  journal={The International Journal of Robotics Research},
  year={2023}
}

@article{zhu2025objectvla,
  title={ObjectVLA: End-to-End Open-World Object Manipulation Without Demonstration},
  author={Zhu, Minjie and Zhu, Yichen and Li, Jinming and others},
  journal={arXiv preprint arXiv:2502.19250},
  year={2025}
}

@inproceedings{o2024open,
  title={Open x-embodiment: Robotic learning datasets and rt-x models: Open x-embodiment collaboration 0},
  author={O’Neill, Abby and   others},
  booktitle={ICRA},
  year={2024},
  organization={IEEE}
}

@article{li2024cogact,
  title={Cogact: A foundational vision-language-action model for synergizing cognition and action in robotic manipulation},
  author={Li, Qixiu and  others},
  journal={arXiv:2411.19650},
  year={2024}
}

@article{brohan2022rt,
  title={Rt-1: Robotics transformer for real-world control at scale},
  author={Brohan, Anthony and Brown, Noah and Carbajal, Justice and   others},
  journal={arXiv preprint arXiv:2212.06817},
  year={2022}
}

@inproceedings{karamcheti2024prismatic,
  title={Prismatic vlms: Investigating the design space of visually-conditioned language models},
  author={Karamcheti, Siddharth and Nair, Suraj and others},
  booktitle={ICML},
  year={2024}
}

@article{hu2022lora,
  title={Lora: Low-rank adaptation of large language models.},
  author={Hu, Edward J and Shen, Yelong and Wallis, Phillip and  others},
  journal={ICLR},
  volume={1},
  number={2},
  pages={3},
  year={2022}
}

@article{team2024octo,
  title={Octo: An open-source generalist robot policy},
  author={Team, Octo Model and Ghosh, Dibya and Walke, Homer and  others},
  journal={arXiv preprint arXiv:2405.12213},
  year={2024}
}

@inproceedings{shentu2024llms,
  title={From llms to actions: Latent codes as bridges in hierarchical robot control},
  author={Shentu, Yide and Wu, Philipp and others},
  booktitle={IROS},
  year={2024},
  organization={IEEE}
}

@article{zheng2024tracevla,
  title={Tracevla: Visual trace prompting enhances spatial-temporal awareness for generalist robotic policies},
  author={Zheng, Ruijie and Liang, Yongyuan  and others},
  journal={arXiv:2412.10345},
  year={2024}
}

@inproceedings{walke2023bridgedata,
  title={Bridgedata v2: A dataset for robot learning at scale},
  author={Walke, Homer Rich and Black, Kevin and Zhao, Tony Z and  others},
  booktitle={Conference on Robot Learning},
  pages={1723--1736},
  year={2023},
  organization={PMLR}
}

@article{li2024towards,
  title={Towards generalist robot policies: What matters in building vision-language-action models},
  author={Li, Xinghang and Li, Peiyan and Liu, Minghuan and  others},
  journal={arXiv preprint arXiv:2412.14058},
  year={2024}
}

@article{li2023vision,
  title     = {Vision-Language Foundation Models as Effective Robot Imitators},
  author    = {Li, Xinghang and Liu, Minghuan and Zhang, Hanbo and others},
  journal={arXiv preprint arXiv:2311.01378},
  year={2023}
}

@misc{brohan2023rt2visionlanguageactionmodelstransfer,
      title={RT-2: Vision-Language-Action Models Transfer Web Knowledge to Robotic Control}, 
      author={Anthony Brohan and Noah Brown and Justice Carbajal and others},
      year={2023},
      eprint={2307.15818},
      archivePrefix={arXiv},
      primaryClass={cs.RO}
}

@article{black2024pi_0,
  title={$ \pi_0 $: A Vision-Language-Action Flow Model for General Robot Control},
  author={Black, Kevin and Brown, Noah and Driess, Danny and   others},
  journal={arXiv preprint arXiv:2410.24164},
  year={2024}
}

@article{beyer2024paligemma,
  title={Paligemma: A versatile 3b vlm for transfer},
  author={Beyer, Lucas and Steiner, Andreas and Pinto, Andr{\'e} Susano and  others},
  journal={arXiv preprint arXiv:2407.07726},
  year={2024}
}

@article{kim2025fine,
  title={Fine-tuning vision-language-action models: Optimizing speed and success},
  author={Kim, Moo Jin and Finn, Chelsea and Liang, Percy},
  journal={arXiv preprint arXiv:2502.19645},
  year={2025}
}

@article{pertsch2025fast,
  title={Fast: Efficient action tokenization for vision-language-action models},
  author={Pertsch, Karl and Stachowicz, Kyle and Ichter, Brian and others},
  journal={arXiv preprint arXiv:2501.09747},
  year={2025}
}

@article{chen2023pali,
  title={Pali-x: On scaling up a multilingual vision and language model},
  author={Chen, Xi and Djolonga, Josip and Padlewski, Piotr and   others},
  journal={arXiv preprint arXiv:2305.18565},
  year={2023}
}

@article{driess2023palm,
  title={Palm-e: An embodied multimodal language model},
  author={Driess, Danny and Xia, Fei and Sajjadi, Mehdi SM and   others},
 journal={arXiv preprint arXiv:2303.03378},
  year={2023},

}

@misc{Orin_specification,
  title={{NVIDIA Jetson AGX Orin Series Technical Brief: a giant leap forward for robotics and edge AI applications}},
  author={Leela, S Karumbunathan},
  volume={4},
  year={2022}
}

@article{li2025hamster,
  title={Hamster: Hierarchical action models for open-world robot manipulation},
  author={Li, Yi and Deng, Yuquan and  others},
  journal={arXiv:2502.05485},
  year={2025}
}

@inproceedings{zhan-etal-2024-rethinking-token,
    title = {Rethinking Token Reduction for State Space Models},
    author = {Zheng Zhan and Yushu Wu and Zhenglun Kong  and others},
    booktitle = {EMNLP},
    month = {nov},
    year = {2024},
    publisher = {ACL}
}

@inproceedings{
zhan2024exploring,
title={Exploring Token Pruning in Vision State Space Models},
author={Zheng Zhan and Zhenglun Kong and Yifan Gong and others},
booktitle={NeurIPS},
year={2024},
}

@inproceedings{
shen2025sparse,
title={Sparse Learning for State Space Models on Mobile},
author={Xuan Shen and Hangyu Zheng and Yifan Gong and others},
booktitle={ICLR},
year={2025}
}

@article{mi2026effective,
  title={Effective MoE-based LLM Compression by Exploiting Heterogeneous Inter-Group Experts Routing Frequency and Information Density},
  author={Mi, Zhendong and Chen, Yixiao and Zhao, Pu and others},
  journal={arXiv preprint arXiv:2602.09316},
  year={2026}
}

@inproceedings{shen2024search,
 author = {Shen, Xuan and Zhao, Pu and Gong, Yifan and others},
 booktitle = {Advances in Neural Information Processing Systems},
 title = {Search for Efficient Large Language Models},
 volume = {37},
 year = {2024}
}

@article{shen2025efficient,
  title={Efficient Reasoning with Hidden Thinking},
  author={Shen, Xuan and Wang, Yizhou and others},
  journal={arXiv preprint arXiv:2501.19201},
  year={2025}
}

@inproceedings{
shen2025fastcar,
title={Fastcar: Cache Attentive Replay for Fast Auto-Regressive Video Generation on the Edge},
author={Xuan Shen and Weize Ma and Yufa Zhou   and others},
booktitle={ICLR},
year={2026}
}

@article{shen2025draftattention,
  title={DraftAttention: Fast Video Diffusion via Low-Resolution Attention Guidance},
  author={Shen, Xuan and Han, Chenxia and Zhou, Yufa and others},
  journal={arXiv preprint arXiv:2505.14708},
  year={2025}
}

@inproceedings{
zhan2024fast,
 title = {Fast and Memory-Efficient Video Diffusion Using Streamlined Inference},
 author={Zheng Zhan and Yushu Wu and Yifan Gong and others},
 volume = {37},
 year = {2024}
}

@article{yang2026alter,
  title={Alter: All-in-one layer pruning and temporal expert routing for efficient diffusion generation},
  author={Yang, Xiaomeng and Lu, Lei and Fan, Qihui and Yang, Changdi and Lin, Juyi and Wang, Yanzhi and Zhang, Xuan and Gao, Shangqian},
  journal={Advances in Neural Information Processing Systems},
  volume={38},
  pages={128571--128599},
  year={2026}
}

@article{taherin2025cross,
  title={Cross-Platform Scaling of Vision-Language-Action Models from Edge to Cloud GPUs},
  author={Taherin, Amir and Lin, Juyi and Akbari, Arash and Akbari, Arman and Zhao, Pu and Chen, Weiwei and Kaeli, David and Wang, Yanzhi},
  journal={arXiv preprint arXiv:2509.11480},
  year={2025}
}

@INPROCEEDINGS{shen2025quartdepth,
  title={QuartDepth: Post-Training Quantization for Real-Time Depth Estimation on the Edge},
  author={Shen, Xuan and Ma, Weize and Liu, Jing and others},
  booktitle={CVPR}, 
  year={2025}
}

@inproceedings{yang2023pruning,
  title={Pruning parameterization with bi-level optimization for efficient semantic segmentation on the edge},
  author={Yang, Changdi and Zhao, Pu and Li, Yanyu and others},
  booktitle={CVPR},
  year={2023}
}

@inproceedings{zhao-etal-2024-pruning,
    title = "Pruning Foundation Models for High Accuracy without Retraining",
    author = "Pu Zhao and Fei Sun and Xuan Shen and others",
    booktitle = "Findings of EMNLP 2024",
    month = nov,
    year = "2024",
    publisher = "ACL",
    pages = "9681--9694",
}

@article{shen2024lazydit, title={LazyDiT: Lazy Learning for the Acceleration of Diffusion Transformers}, volume={39},  number={19}, journal={AAAI}, author={Shen, Xuan and Song, Zhao and Zhou, Yufa and   others}, year={2025}, month={Apr.}  }

@article{shen2024numerical, title={Numerical Pruning for Efficient Autoregressive Models}, volume={39},  number={19}, journal={AAAI}, author={Shen, Xuan and Song, Zhao and Zhou, Yufa and   others}, year={2025}, month={Apr.} }

@article{zhao2025open,
  title={Open-Source Multimodal Moxin Models with Moxin-VLM and Moxin-VLA},
  author={Zhao, Pu and Akbari, Arash and Shen, Xuan and   others},
  journal={arXiv preprint arXiv:2512.22208},
  year={2025}
}
%




\end{document}